\documentclass[conference]{IEEEtran}
\IEEEoverridecommandlockouts

\usepackage{cite}
\usepackage{amsmath,amssymb,amsfonts}
\usepackage{algorithmic}
\usepackage{textcomp}
\usepackage{url}
\usepackage{booktabs}
\usepackage{array}
\usepackage{multirow}
\usepackage{arydshln}
\usepackage{graphicx}  
\usepackage{siunitx}
\def\BibTeX{{\rm B\kern-.05em{\sc i\kern-.025em b}\kern-.08em T\kern-.1667em\lower.7ex\hbox{E}\kern-.125emX}}

\begin{document}

\title{AsyncSwitch: Asynchronous Text-Speech Adaptation for Code-Switched ASR}

\author{\IEEEauthorblockN{1\textsuperscript{st} Tuan Nguyen}
\IEEEauthorblockA{\textit{Institute for Infocomm Research (I²R)} \\
\textit{A*STAR}\\
Singapore \\
nguyenvat@i2r.a-star.edu.sg}
\and
\IEEEauthorblockN{2\textsuperscript{nd} Huy-Dat Tran}
\IEEEauthorblockA{\textit{Institute for Infocomm Research (I²R)} \\
\textit{A*STAR}\\
Singapore \\
hdtran@i2r.a-star.edu.sg}}

\maketitle

\begin{abstract}
Developing code‐switched ASR systems is challenging due to language ambiguity and limited exposure to multilingual, code‐switched data, while collecting such speech is costly. Prior work generates synthetic audio from text, but these methods are computationally intensive and hard to scale. We introduce AsyncSwitch, a novel asynchronous adaptation framework that leverages large‐scale, text‐rich web data to pre‐expose ASR models to diverse code‐switched domains before fine‐tuning on paired speech–text corpora. Our three‐stage process (1) trains decoder self‐attention and feedforward layers on code‐switched text, (2) aligns decoder and encoder via cross‐attention using limited speech–text data, and (3) fully fine‐tunes the entire model. Experiments with Whisper on Malay–English code-switching demonstrate a 9.02\% relative WER reduction, while improving monolingual performance in Singlish, Malay, and other English variants.

\end{abstract}


\begin{figure*}[t!] 
    \centering
    \resizebox{\textwidth}{!}{%
    \includegraphics[width=\textwidth]{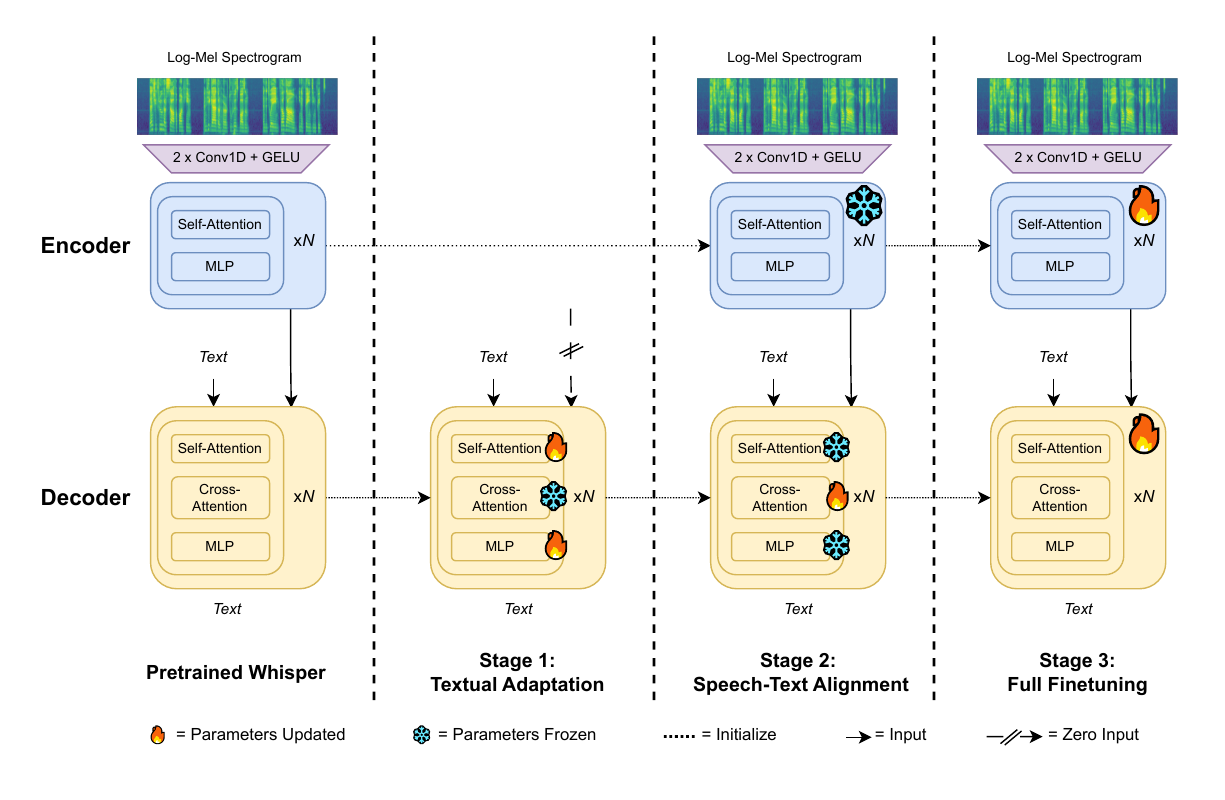} 
    }
    \caption{Overview of AsyncSwitch: a three-stage asynchronous adaptation framework using large-scale text and speech-text data for code-switched ASR}
    \label{fig:textual_figure} 
\end{figure*}

\section{Introduction}

Code-switching—switching between languages within the same conversation—is a common and natural way of speaking in many multilingual communities. This is especially true in Southeast Asia, where people often mix their native language with English in everyday conversations~\cite{wikipediaMalay}. However, this kind of speech remains a major challenge for Automatic Speech Recognition (ASR) systems, and even powerful models like Whisper~\cite{whisper_paper} perform poorly on mixed-language input. One key reason is the imbalance in training data: about two-thirds of the data is in English~\cite{whisper_paper}, while the remaining 99+ languages~\cite{whisper_paper,winata-etal-2020-meta} have much less coverage. For instance, Malay has fewer than 100 hours of training data \cite{whisper_paper}, and code-switched speech is even more limited. Because of this gap, models struggle to learn and accurately recognize speech that mixes multiple languages.

Multilingual environments like Malay and Singapore create fluid code-switching patterns that current ASR systems struggle to handle. This challenge is further worsened by limited code-switched labeled data for regional languages. Conventional finetune approaches \cite{semi_sup_asr_low_res}  often face language ambiguity, and phoneme confusion \cite{whisper_lm} due to insufficient domain coverage, lack of diversity, or bias toward dominant languages—leading to misrecognitions \cite{language_bias_slt_24} or hallucinated outputs \cite{whisper_hallu}.


Recent work by Tuan et al.~\cite{phrase_mix_paper} addressed this issue by generating synthetic code-switched speech using a phrase mixing method, enabling effective fine-tuning of multilingual ASR models. While their approach demonstrated strong performance gains, it required costly and computationally intensive speech generation pipelines.

We observed that large-scale pretrained ASR models like Whisper demonstrate strong acoustic capabilities, thanks to training on millions of hours of speech data~\cite{whisper_paper,seamless2023}. However, they still struggle with code-switching transcription. A key reason is their reliance on paired speech-text data, which limits language understanding—especially for underrepresented and mixed-language inputs. This raises a central question: Can abundant textual data help compensate for the lack of large-scale speech-text resources and improve code-switching performance in pretrained ASR models?

To address this gap, we propose AsyncSwitch, a novel \textbf{Asynch}ronous adaptation framework explicitly designed to improve ASR performance in Code-\textbf{Switch}ing scenarios, while also benefiting monolingual low-resource settings. To overcome the decoder's limited understanding of code-switched language, we introduce a three-stage adaptation process: (1) training the decoder’s self-attention and feedforward layers on target code-switched text, (2) aligning the decoder and encoder via cross-attention using a small amount of speech–text data, and (3) fully fine-tuning the entire model. This work is inspired by recent advances in the Large Speech Language Model (SLM) paradigm~\cite{speech_llm_survey}, which highlight the potential of large-scale textual pretraining followed by audio adaptation for multilingual and low-resource ASR tasks.

Our contributions are as follows.
\begin{itemize}
    \item We propose \textbf{AsyncSwitch}, a novel three-stage asynchronous ASR adaptation framework that leverages abundant textual data and limited bilingual, and code-switched speech for improved code-switching performance.
    \item We achieve significant WER reductions on Bahasa Malay and Singlish datasets: 9.02\% for Malay-English code-switching~\cite{phrase_mix_paper}, 17.35\% for monolingual Malay, and 14.5\% for Singlish.
    \item Our method outperforms commercial ASR systems (I\textsuperscript{2}R A*STAR~\cite{i2r_astar_api} and Azure~\cite{azure_api}) by 7.7\% and 11.9\% respectively.
    \item Our method prevents catastrophic forgetting and improves performance on the Open ASR Leaderboard's diverse English scenarios~\cite{open-asr-leaderboard}.
\end{itemize}




By prioritizing the challenges of code-switching in low-resource multilingual environments—and doing so without heavily depending on synthetic speech—this work contributes to more inclusive, efficient, and adaptable ASR systems.

The remainder of this paper is organized as follows: Section \ref{sec:related_work} reviews related work on text-based adaptation and code-switching ASR. Section \ref{sec:method} details the proposed three-stage method. Section \ref{sec:experiments} outlines the experimental settings, including datasets and training configurations. Section \ref{sec:results} presents the results and analysis. Section \ref{sec:limitation} discusses our limitations, and Section \ref{sec:conclusion} concludes with a summary and future directions.

\section{Related Work}
\label{sec:related_work}

\subsection{Text-Based and Internal LM Adaptation}

Recent efforts have explored leveraging unpaired text data to enhance ASR performance, especially in low-resource settings. USTR-CT~\cite{ustr_ct_is23} proposed a temporary text encoder to adapt Conformer-Transducer models using text-only data, achieving strong gains without increasing inference complexity. Meanwhile, Radford et al.~\cite{whisper_paper} demonstrated Whisper’s robustness across multilingual and multitask settings, though primarily relying on paired data.

Internal language model (LM) adaptation has also gained attention. ILME-based methods~\cite{ilme_asr_slt_2021} subtract estimated internal LM scores during decoding to better integrate external LMs, while AdaBERT-CTC~\cite{vuong-etal-2023-adabert} fine-tunes a BERT-style encoder on domain-specific text. However, these approaches typically focus on inference-time integration or domain adaptation, and often neglect multilingual or code-switched challenges.

In contrast, our work introduces a three-stage pipeline that begins with internal LM adaptation on unpaired text—including code-switched data—followed by cross-attention alignment and end-to-end fine-tuning. This design enables systematic exploitation of large-scale text corpora and addresses both data scarcity and code-switching robustness in a unified manner.

\subsection{Code-Switching ASR}

Code-switching remains a persistent challenge in ASR, requiring systems to handle dynamic language transitions. Prior studies addressed this using unified modeling~\cite{ylmaz18c_interspeech}, syntactic priors~\cite{chi-bell-2022-improving}, or synthetic augmentation~\cite{sharma20c_interspeech}. However, many of these rely on annotated code-switched corpora, which are rare in under-resourced settings.

Tuan et al.~\cite{phrase_mix_paper} proposed a scalable alternative by generating 3,000 hours of synthetic phrase-mixed CS speech across Malay-English, Mandarin-Malay, and Tamil-English. Their method—based on translation, alignment, and audio splicing—achieved notable improvements when fine-tuning large ASR models like Whisper, SeamlessM4T, and MMS. They also introduced benchmark test sets for CS, Singlish, and monolingual ASR, which we adopt in our evaluation.


We build on Tuan et al.'s [5] insight that synthetic speech can benefit model training, and we directly use a portion (20\%) of their synthetic dataset in our speech-text tuning setup. While our approach includes audio splicing as part of the data preparation process, it is not the primary focus. Instead, our method emphasizes adapting Whisper using predominantly large-scale text supervision.

\subsection{Low-Resource ASR}

Low-resource ASR research often centers on transfer learning~\cite{sharma20c_interspeech}, self-supervised learning~\cite{SpeechLM}, or synthetic speech from TTS~\cite{tts_cs_asr}. While these approaches reduce dependence on labeled data, they often require either large unlabelled audio corpora or high-quality TTS systems—both limiting factors in many languages.

Our approach addresses this by front-loading adaptation into the language modeling component using only text. By priming the decoder with code-switched and monolingual text before aligning with speech, we achieve robust performance even in low-resource scenarios like Bahasa Malay and Singlish, without large-scale audio resources.

\section{Proposed Method: AsyncSwitch}
\label{sec:method}
This section details our methodology for adapting the pretrained Whisper ASR model \cite{whisper_paper} to improve performance on low-resource languages like Malay through a three-stage approach. Figure \ref{fig:textual_figure} illustrates the details of our method.

\subsection{Pretrained ASR Model: Whisper}
We utilize the Whisper-Large-v3\footnote{https://huggingface.co/openai/whisper-large-v3} model \cite{whisper_paper}, trained on 5 million hours of diverse, weakly supervised audio data covering multiple languages and tasks. Despite its multilingual capabilities, adaptation is often needed for optimal performance on specific low-resource languages or domains not heavily represented in the initial training data.

\subsection{AsyncSwitch: An Asynchronous Text-Speech Adaptation Framework}

We adapt a pretrained Whisper encoder–decoder model \(\bigl(\theta = \{\theta_{E},\,\theta_{D}\}\bigr)\) to a low-resource target language via three successive stages.  Let $x\in\mathbb{R}^{L\times d} \quad\text{(audio encoder features)}$, $y=(y_{1},\dots,y_{T})  \quad\text{(target token sequence)}$, 
and decompose the decoder parameters as
\[
\theta_{D} = \{\theta_{\mathrm{SA}},\,\theta_{\mathrm{CA}},\,\theta_{\mathrm{FF}},\,\theta_{\mathrm{out}}\},
\]
where \(\theta_{\mathrm{SA}}\), \(\theta_{\mathrm{CA}}\), and \(\theta_{\mathrm{FF}}\) denote the self-attention, cross-attention, and feed-forward blocks in each layer, and \(\theta_{\mathrm{out}}\) is the final projection to the vocabulary.

\subsubsection{Stage 1: Decoder Internal LM Adaptation}
We zero out the encoder output (\(x=0\)) so that the decoder functions purely as a conditional language model.  We update only 
\(\theta_{\mathrm{SA}},\,\theta_{\mathrm{FF}},\,\theta_{\mathrm{out}}\) 
(and keep \(\theta_{\mathrm{CA}}\) frozen) to learn domain text patterns via next-token cross-entropy:
\[
\mathcal{L}_{1}
= -\sum_{t=1}^{T}\log p_{\theta_{\mathrm{SA}},\,\theta_{\mathrm{FF}},\,\theta_{\mathrm{out}}}\bigl(y_{t}\mid y_{<t},\,x=0\bigr).
\]
This stage leverages large unlabeled text corpora to adapt the internal LM without disturbing audio–text alignment.

\subsubsection{Stage 2: Speech–Text Alignment Fine-tuning}
We re-activate the acoustic encoder (\(\theta_{E}\)) (but still freezing) and unfreeze only the decoder’s cross-attention \(\theta_{\mathrm{CA}}\), holding \(\{\theta_{E},\,\theta_{\mathrm{SA}},\,\theta_{\mathrm{FF}},\,\theta_{\mathrm{out}}\}\) fixed. Using paired dataset \((x,y)\), we optimize
\[
\mathcal{L}_{2}
= -\sum_{t=1}^{T}\log p_{\theta_{\mathrm{CA}}}\bigl(y_{t}\mid y_{<t},\,x\bigr),
\]
thereby strengthening the model’s ability to align encoder representations to the newly adapted decoder LM.

\subsubsection{Stage 3: Full End-to-End Fine-tuning}
Finally, we unfreeze all parameters \(\{\theta_{E},\,\theta_{\mathrm{SA}},\,\theta_{\mathrm{CA}},\,\theta_{\mathrm{FF}},\,\theta_{\mathrm{out}}\}\) and fine-tune end-to-end on the same paired data:
\[
\mathcal{L}_{3}
= -\sum_{t=1}^{T}\log p_{\theta}\bigl(y_{t}\mid y_{<t},\,x\bigr).
\]
This global optimization refines both acoustic and linguistic components to the target domain.

\vspace{1ex}
\noindent The progression from text-only LM adaptation (Stage 1) through targeted alignment (Stage 2) to full fine-tuning (Stage 3) provides a balanced trade-off between data efficiency and modeling flexibility (see Figure \ref{fig:textual_figure}). Without first two stages, the model cannot adapt to text only data, which is very valuable in context of low-resources or domain-adaptation.

\begin{table*}
    \centering
    \caption{Evaluation Results on Singlish, Malay, and Code-Switched Datasets. \textit{Italic values indicate group-wise averages.}}
    \renewcommand{\arraystretch}{1.12} 
    \resizebox{\textwidth}{!}{%
    {\LARGE
    \begin{tabular}{l cc|rrr rrr rrr c}
    \toprule
    \multirow{2}{*}{\textbf{Model}} & 
    \multirow{2}{*}{\textbf{\#(B)}} & 
    \multirow{2}{*}{\shortstack{\textbf{OpenASR} \cite{open-asr-leaderboard} \\ \textbf{Leaderboard}}} & 
    \multicolumn{3}{c}{\textbf{Singlish}} & 
    \multicolumn{3}{c}{\textbf{Malay}} & 
    \multicolumn{3}{c}{\textbf{CodeSwitch Malay-English}} & 
    \multirow{2}{*}{\textbf{Avg.}} \\
    \cmidrule(r){4-6} \cmidrule(r){7-9} \cmidrule(r){10-12}
     & & & NLB \cite{nlb_dataset} & IMDA3 \cite{imda_nsc} & \textit{Avg. EN} & Noisy \cite{phrase_mix_paper} & Convo \cite{malayconvo2025} & \textit{Avg. BM} & Reading \cite{phrase_mix_paper} & IMDA4 \cite{imda_nsc} & \textit{Avg. CS} & \\
    \midrule
    \textbf{Baseline} \\
    \textsc{Whisper-Orig.} \cite{whisper_paper}  & 1.6  & 7.44 & 22.52 & 18.61 & \textit{20.57} & \underline{28.06} & 31.57 & \textit{29.82} & 7.06 & 34.82 & \textit{20.94} & \textit{23.77} \\
    \textsc{Whisper-5k} & 1.6 & \underline{7.27} & 21.68 & 15.84 & \textit{18.76} & 31.60 & 20.61 & \textit{26.11} & \underline{5.92} & \underline{31.54} & \underline{\textit{18.73}} & \textit{21.20} \\
    \quad + LM 38M & - & - & \underline{21.30} & \underline{14.53} & \underline{\textit{17.92}} & 32.04 & \underline{19.85} & \underline{\textit{25.95}} & 7.50 & 31.70 & \textit{19.60} & \underline{\textit{21.15}} \\
    \midrule
    \textbf{SpeechLLM} \\
    \textsc{\shortstack{MERaLiON-AudioLLM\\-Whisper-SEA-LION}} \cite{meralion_model} & 9.9 & - & 25.51 & 13.17 & \textit{19.34} & 52.06 & 47.64 & \textit{49.85} & 15.74 & 35.50 & \textit{25.62} & \textit{31.60} \\
    \midrule
    \textbf{Commercial-Grade ASR} \\
     \textnormal{I\textsuperscript{2}R, A*STAR} \cite{i2r_astar_api} & - & 8.92 & 21.06 & 12.71 & \textit{16.89} & 25.58 & 23.43 & \textit{24.51} & \underline{5.94} & \underline{29.69} & \underline{\textit{17.82}} & \underline{\textit{19.74}} \\
    \textsc{Azure ASR API} \cite{azure_api} & -  & - & \underline{17.66} & \textbf{8.68} & \textbf{\textit{13.17}} & \underline{23.43} & \underline{20.92} & \underline{\textit{22.18}} & 22.92 & 30.47 & \textit{26.70} & \textit{20.68} \\
    \midrule
    \textbf{Phrase-Mixed CS EN-ZH-BM-TA} \\
    \textsc{WhisperTurbo-V3} \cite{phrase_mix_paper} & 0.8 & - & 45.01 & \underline{9.29} & \textit{27.15} & 33.06 & \textbf{12.88} & \textit{22.97} & 16.75 & \textbf{27.80} & \underline{\textit{22.28}} & \textit{24.13} \\
    \textsc{MMS-1B-All} (FT w/ BPE) \cite{phrase_mix_paper} & 1.0 & - &  33.41 & 19.83 & \textit{26.62} & 36.31 & 22.48 & \textit{29.40} & 23.18 & 40.15 & \textit{31.67} & \textit{29.23}  \\
    \textsc{SeamlessM4T-v2} \cite{phrase_mix_paper}  & 1.6  & - &  \underline{27.24} & 12.16 & \underline{\textit{19.70}} & \underline{27.46} & 15.71 & \underline{\textit{21.59}} & \underline{14.58} & 30.26 & \textit{22.42} & \underline{\textit{21.24}} \\
    \midrule
    \textbf{AsyncSwitch (Our)} \\
    \textsc{Whisper-38M-5k} & 1.6 & \textbf{7.04} & 20.69 & 11.39 & \textit{16.04} & \textbf{22.75} & 20.41 & \textit{\textbf{21.58}} & \textbf{5.10} & 28.98 & \textit{\textbf{17.04}} & \textbf{18.22} \\
    \bottomrule
    \end{tabular}
    }
    }
    \label{tab:results}
\end{table*}

\section{Experimental Settings}
\label{sec:experiments}
\subsection{Dataset}
\subsubsection{Textual Data (Stage 1)}

We compiled approximately 38.3 million (\textsc{38M}) text utterances from the following sources:

\begin{itemize}
    \item \textbf{Sealion Pile BM corpus} \cite{aisingapore2023seapile}: 29.2 million Malay utterances.
    \item \textbf{IMDA NSC text subsets \{1,2,3,5,6\}} \cite{imda_nsc}: 7.1 million Singlish utterances, covering prompted readings, debates, finance, phone calls, and more.
    \item \textbf{LibriSpeech texts} \cite{librispeech}: We used 280k US English Text from LibriSpeech.
    \item \textbf{Filtered Malay YouTube transcripts}\footnote{https://huggingface.co/datasets/mesolitica/pseudolabel-malaysian-youtube-whisper-large-v3}: 1.7 million utterances. Our processing included: (1) filtering out speech not in the target language using ECAPA-TDNN \cite{speechbrain2021langid_ecapa_tdnn}; (2) filtering text using FastText language detection \cite{facebook2020fasttextlid}; (3) removing utterances with repeated 2- or 3-grams occurring more than 4 times \cite{filter_ngram}; and (4) excluding utterances with fewer than 32 tokens. This resulted in 14,000 hours of Malay YouTube speech and 1.7 million text utterances.
\end{itemize}

\subsubsection{Speech Data (Baseline Training \& Stages 2/3)}
We used a combination of 1k-hours English, included: Singlish IMDA NSC \{1,2,5,6\} - each 180 hours \cite{imda_nsc} and US English LibriSpeech 250 hours \cite{librispeech}, 1k-hours Malay \cite{phrase_mix_paper}, 1k-hours phrase-mixed \cite{phrase_mix_paper} - all are same as \cite{phrase_mix_paper}, and 2k-hours new added, sampled from Filtered Malay YouTube.


\subsection{Training Settings}

\subsubsection{Comparison} 

\textbf{Baseline.} We compared our proposed model against the original Whisper-Large-V3\cite{whisper_paper} (\textsc{Whisper-Orig.}) and a version fine-tuned on a 5k-hour dataset (\textsc{Whisper-5k}). We also evaluated commercial-grade ASR systems, including Azure Speech-to-Text \cite{azure_api} and the \textnormal{I\textsuperscript{2}R, A*STAR} ASR API \cite{i2r_astar_api}, both optimized for Southeast Asian languages.

\textbf{With Language Model.} For \textsc{Whisper-5k}, we applied a 5-gram language model trained on 38M tokens with beam size 2, following \cite{whisper_lm}. We tuned hyperparameters $\alpha \in [0,0.1]$ and $\beta \in [-0.2,0.2]$ on 500 samples from the 5k-hour dataset (excluding 1k hours of phrase-mixed data) across 4 trials.

\textbf{SpeechLLM.} We conducted comparative evaluations against established SpeechLLMs \cite{speech_llm_survey}, with particular attention to Southeast Asian variants, including \textsc{MERaLiON-AudioLLM-Whisper-SEA-LION}\cite{meralion_model}. This model is tailored for Southeast Asian languages through training on SealionPile \cite{aisingapore2023seapile} (our Malay text comprises a subset of this dataset, representing approximately 0.29\% of total tokens) and large-scale speech-instruction tuning data that includes code-switched Singapore languages.

\subsubsection{Three-Stage Adaptation}

\begin{itemize}
    \item \textbf{Stage 1 - Textual Adaptation:} The decoder is trained on \textsc{38M} text utterances with a peak learning rate of 2e-5, 10\% linear warm-up, cosine decay, and a batch size of 128.
    \item \textbf{Stage 2 - Speech-Text Alignment:} Cross-attention layers are trained on 5k hours of paired speech-text data for 1 epoch.
    \item \textbf{Stage 3 - Full Fine-tuning:} The entire model is fine-tuned on the same 5k-hours speech-text dataset for 2 epochs (45k updates), resulting in the final model, \textsc{Whisper-38M-5k}.
\end{itemize}

All speech-text experiments (baseline, Stage 2, and Stage 3) used a peak learning rate of 2e-5, 20\% linear warm-up, cosine decay, and a batch size of 32. Default SpecAugment settings \cite{spec_aug} were applied, along with noise augmentation using the Noise subset of Musan \cite{musan2015} at \{10, 40\} dB.

During training, we used language prompting \cite{whisper_paper} by prepending either the \texttt{<|ms|>} (Malay) or \texttt{<|en|>} (English) token to each utterance, based on the dominant language (determined by word count).

After fine-tuning, all models were merged with the original \textsc{Whisper-Large-v3} using a merging ratio of 0.4 (with the original model contributing 0.6). Detailed merging ratios are provided in Section \ref{tab:merging_ratio}.

All experiments were conducted on four A40 GPUs (44 GB each), using DeepSpeed ZeRO-3 optimization.

\subsection{Evaluation Settings}

We evaluate the fine-tuned model on code-switching Malay-English (BM-EN), Singlish (EN), and Malay (BM) scenarios, under noisy and conversational conditions, with a focus on low-resource domains:

\begin{itemize}
\item \textbf{CS BM-EN:} ChatGPT-generated conversations, the Singaporean Reading test set \cite{phrase_mix_paper}, and IMDA4 BM-EN \cite{imda_nsc}.
\item \textbf{Singlish:} Noisy historical interviews (past-century) (NLB) \cite{nlb_dataset} and IMDA3 Conversation \cite{imda_nsc}.
\item \textbf{Malay:} Conversational and noisy sets from \cite{malayconvo2025, phrase_mix_paper}.
\end{itemize}

We used the combined \texttt{<|ms|><|en|>} prompt~\cite{adapt_whisper_cs_en_zh} across all test sets to support code-switching output.

The model was also evaluated on the OpenASR Leaderboard \cite{open-asr-leaderboard} (English), to assess catastrophic forgetting.

Additionally, we used the Code-Mixing Index (CMI) \cite{cmi_value} to quantify code-switching in text corpora. Higher CMI values indicate more code-switching patterns.

\section{Results}
\label{sec:results}

\subsection{Main Results}







\textbf{Baseline Comparisons.} \textsc{Whisper-38M-5K} demonstrates substantial performance gains, exceeding the original Whisper model by 23.35\% and models trained on equivalent labeled speech data (\textsc{Whisper-5k}) by 14.05\%. The largest improvements are observed on the Singlish IMDA3 and Malay Noisy datasets. While external language models trained on the same text data provide a marginal 0.23\% improvement, they consistently underperform our proposed method.

\textbf{Commercial Systems.} Our method achieves 7.7\% relative improvement over the \textnormal{I\textsuperscript{2}R, A*STAR} ASR \cite{i2r_astar_api} across all test sets. While Azure \cite{azure_api} performs better on Singlishs, it underperforms significantly on Malay and Code-Switch Malay-English.

\textbf{SpeechLLM.} Our method outperforms \textsc{MERaLiON-AudioLLM-Whisper-SEA-LION} \cite{meralion_model} by 42.34\% overall while being $6\times$ smaller in size.

\textbf{Large-Scale Code-Switching Models}. \textsc{WhisperTurbo-V3} EN-ZH-BM-TA performs best on Singlish IMDA3, Malay Convo, and Code-Switched IMDA4, but fails on Code-Switch reading (16.75 vs 5.10 for our model). Overall, our method achieves 24.49\%, 37.67\%, and 14.22\% improvements over \textsc{WhisperTurbo-V3}, \textsc{MMS-1B-All}, and \textsc{SeamlessM4T-v2}, respectively.

\textbf{Catastrophic Forgetting.} Our method improves English speech recognition by 5.37\% compared to the original model, demonstrating successful knowledge retention without degradation.

These results clearly state that AsyncSwitch demonstrates strong performance on code-switching tasks while maintaining the same model structure. The method shows consistent improvements across baselines, with the largest gains on code-switched BM-EN datasets, providing an effective approach for code-switch speech recognition.

\subsection{Ablation Study}

\subsubsection{Results on different stage of training}

Table \ref{tab:training_stages} shows that our three-stage AsyncSwitch training provides incremental improvements. Stages 1-2 with domain text and minimal supervised speech data substantially improve Singlish and surpass \textsc{Whisper-5k}, but show limited improvement for low-resource Malay and minimal code-switch gains (gap with Original to 1.2\% relatively). Stage 3 with full supervision achieves optimal performance across scenarios, confirming that early domain text incorporation provides a foundation while comprehensive fine-tuning is essential for all domains.

\begin{table}[h] 
    \centering
    \caption{Evaluation Results at Different Training Stages} 
    \resizebox{\columnwidth}{!}{%
    \begin{tabular}{lcccc} 
    \toprule
    \textbf{Training Stages} & \textbf{Avg. EN} & \textbf{Avg. BM} & \textbf{Avg. CS} & \textbf{Avg.} \\ 
    \midrule
    \textsc{Whisper-Orig.}               & 20.57   & 29.82   & 20.94   & 23.77      \\
    \textsc{Whisper-5k}           & 18.76          & 26.11          & 18.73          & 21.20 \\
    \midrule
    Phase1 - 0.4           & 19.16   & 31.32   & 23.41    & 24.63      \\
    Phase2 - 0.4           & 18.37       & 36.31       & 21.19       & 25.29          \\ 
    Phase3 - 0.4           & \textbf{16.04}   & \textbf{21.58}   & \textbf{17.04}   & \textbf{18.22} \\ 
    \bottomrule
    \end{tabular}
    }
    \label{tab:training_stages} 
\end{table}



\subsubsection{Scale of textual data}
We compared smaller-but-focused code-switch text fine-tuning with MalayYoutube 1.7M text (\textsc{Whisper-1.7M-5k}) against our 38M text approach. Table \ref{tab:scaling_results} shows the smaller text model performs better on Malay and Code-Switch (3.8\% and 3.72\% relatively) but degrades Singlish performance (11.58\% relatively) due to its narrow focus. Overall, \textsc{Whisper-38M-5k} achieves better performance. While the 1.7M MalayYoutube text has high CMI values (Table \ref{tab:dataset_stats}), the significantly larger and more diverse 38M corpus better handles real-world code-switching scenarios.

\begin{table}[h] 
    \centering
    \caption{Comparison of Results by Textual Data Size} 
    \resizebox{\columnwidth}{!}{%
    \begin{tabular}{lcccc} 
    \toprule
    \textbf{Text Increment} & \textbf{Avg. EN} & \textbf{Avg. BM} & \textbf{Avg. CS} & \textbf{Avg.} \\ 
    \midrule
    \textsc{Whisper-Orig.}              & 20.57          & 29.82          & 20.94          & 23.77          \\
    \textsc{Whisper-5k}           & 18.76          & 26.11          & 18.73          & 21.20          \\
    \midrule
    \textsc{Whisper-1.7M-5k}   & 18.14          & \textbf{20.79} & \textbf{16.43} & 18.45          \\
    \textbf{\textsc{Whisper-38M-5k}} & \textbf{16.04} & 21.58          & 17.04          & \textbf{18.22} \\ 
    \bottomrule
    \end{tabular}
    }
    \label{tab:scaling_results} 
\end{table}

\begin{table}[h]
    \centering
    \renewcommand{\arraystretch}{1.1} 
    \caption{Code-Mixed Index (CMI) Statistics Across Datasets}
    \resizebox{\columnwidth}{!}{%
    \begin{tabular}{l l r r}
        \toprule
        \textbf{Type} & \textbf{Name} & \textbf{\#Utt.(M)} & \textbf{CMI} \\
        \midrule
        Speech-Text & Speech-Text 5k hours & 2.8 & 27.96 \\
        \midrule
        \multirow{5}{*}{Text 38M} 
        & SealionPile BM \cite{aisingapore2023seapile} & 29.2 & 31.78 \\
        & MalayYT Filtered & 1.7 & 31.78 \\
        & IMDA\{1,2,3,5,6\} \cite{imda_nsc} & 7.1 & 6.66 \\
        & LibriSpeech \cite{librispeech} & 0.3 & 1.54 \\
        \cmidrule{4-4}
        &  & & 28.41 \\
        \midrule
        Phrase-Mixed & BM-EN 1k hours \cite{phrase_mix_paper} & 1.1 & 38.37 \\
        \midrule 
        \multirow{2}{*}{\shortstack{BM-EN \\ CS Test}} 
        & Reading \cite{phrase_mix_paper} & - & 36.88 \\
        & IMDA4 \cite{imda_nsc} & - & 14.81 \\
        \bottomrule
    \end{tabular}
    }
    \label{tab:dataset_stats}
\end{table}




\subsubsection{Optimal Merging Ratio}

Table \ref{tab:merging_ratio} presents merging ratios using linear interpolation \cite{linear_interpolate_space} between the original Whisper model and our domain-specific fine-tuned model. We choose a merging ratio of $0.4$ for best code-switching performance (17.04) while maintaining acceptable Singlish and Malay results. Although $0.8$ achieves the best overall average (16.77), it degrades performance on all other scenarios (Singlish, CS, OpenASR Leaderboard with diverse English). Higher fine-tuned ratios favor Malay without necessarily improving code-switching.

\begin{table}[h] 
    \centering
    \caption{Evaluation Results for Different Merging Ratios} 
    \resizebox{\columnwidth}{!}{%
    {\large
    \begin{tabular}{lccccc} 
    \toprule
    \textbf{Merging Ratio} & \textbf{OpenASR} \cite{open-asr-leaderboard} & \textbf{EN} & \textbf{BM} & \textbf{CS} & \textbf{Avg.} \\ 
    \midrule
    0.0 \textsc{(Whisper-Orig.)}     &   7.44   & 20.57          & 29.82          & 20.94   & 23.77          \\
    \midrule
    0.2                   & 9.57  & 17.43          & 26.65          & 22.10    & 22.06          \\
    \textbf{0.4 \textsc{(Whisper-38M-5k)}} &  \textbf{7.40}                 & 16.04          & 21.58          & \textbf{17.04}   & 18.22          \\
    0.6                   & 7.57 & \textbf{15.44} & 26.65          & 17.50    & 19.86          \\
    0.8                   & 8.08 & 16.99          & \textbf{15.56} & 17.76    & \textbf{16.77}          \\
    \midrule
    1.0  & 8.94 &  19.34         & 16.01          & 17.95   & 17.76 \\ 
    \bottomrule
    \end{tabular}
    }
    }
    \label{tab:merging_ratio} 
\end{table}

\section{Limitation}
\label{sec:limitation}


This study was conducted within the Malaysian and Singaporean linguistic contexts, where English serves as a prominent language alongside Malay, providing abundant text with high Code-Mixing Index (CMI) values. This unique bilingual environment may limit the generalizability of our findings to regions with different language dynamics or less extensive code-switched text resources. Future applications of AsyncSwitch must carefully evaluate their target text characteristics to ensure domain compatibility with the proposed approach.

\section{Conclusion}
\label{sec:conclusion}

We proposed AsyncSwitch, a novel three-stage asynchronous adaptation framework that leverages large-scale unpaired text and limited speech data to improve ASR in low-resource, code-switched settings. Stage 1 and 2 adapt the decoder language model on domain-specific/code-switched scenarios and align it back with the encoder, while Stage 3 fine-tunes the full model along with merging strategy. Experiments on Malay-English, Malay, and Singlish show that AsyncSwitch outperforms Whisper-Large-v3, SpeechLLM and commercial systems, with additional gains on the OpenASR Leaderboard, demonstrating excellent performance on domain-specific code-switching while improving on English scenarios, showing strong generalization and preventing catastrophic forgetting. The AsyncSwitch framework redefines efficient, scalable solutions for domain-specific, low-resource settings like code-switched ASR. This work establishes a new paradigm where abundant text can unlock high-quality speech recognition for underserved multilingual communities.

\bibliographystyle{IEEEtran}
\bibliography{mybib}

\end{document}